\ifdefined\pdfoutput\pdfoutput=1\fi
\documentclass[pdflatex,sn-nature]{sn-jnl}

\usepackage{graphicx}
\usepackage{amsmath,amssymb}
\usepackage{booktabs}
\usepackage{siunitx}
\usepackage{algorithm}
\usepackage{algpseudocode}
\usepackage{placeins}

\newcommand{\ci}[2]{[\num{#1}, \num{#2}]}%
\begin{document}

\title[A wind memory for odour search]{A wind memory for odour source search in reversing wind}

\author*[1]{\fnm{Junlong} \sur{Shen}}\email{junlong6@ualberta.ca}

\author*[2]{\fnm{Shunsuke} \sur{Shigaki}}\email{shigaki@nii.ac.jp}

\affil[1]{\orgdiv{Department of Electrical and Computer Engineering}, \orgname{University of Alberta}, \orgaddress{\city{Edmonton}, \country{Canada}}}

\affil[2]{\orgdiv{Principles of Informatics Research Division}, \orgname{National Institute of Informatics}, \orgaddress{\city{Tokyo}, \country{Japan}}}

\abstract{After wind reversals, old odour arrives while the sensed wind points away from the source, so
wind followers, which surge upwind, are misled. A hold gate stores the wind at first odour contact
and feeds it to an unchanged follower while the live wind points over \SI{90}{\degree} away. On 60 %
random simulated reversing flows it raised a published moth model's success by 10.8 points (95\% CI %
6.6 to 15.4) and four other followers' by 6.0 to 10.1, but cost the moth model 5 points in steady %
wind. A fixed wind delay was not resolvably worse at insect scale, but in simulation rescaled to a
legged robot's sensors and turn rate the gate beat the best delay for two followers, only one margin %
surviving Bonferroni correction. As the robot never ran the gate and a plain spiral searcher beat the
gated moth model, it is an add-on, not a better searcher.}

\maketitle

\section{Introduction}

A robot searching for a gas leak meets odour as brief filaments separated by clean
air~\cite{murlis1992odor,kowadlo2008robot,ishida2012chemical,chen2019odor,jing2021recent,francis2022gas,reddy2022olfactory}.
Most search strategies borrow the moth's rule of surging upwind while odour is detected and casting
crosswind when it is lost~\cite{kennedy1974pheromone,vickers2000mechanisms,carde2008navigational,vanbreugel2014plume}.
Robots have applied the rule underwater~\cite{farrell2005chemical,li2006moth}, on
wheels~\cite{shigaki2018time,lochmatter2008comparison,harvey2008comparing}, with moth antennae as
sensors~\cite{kuwana1999synthesis} and in insect--machine hybrids driven by a
silkmoth~\cite{ando2013odour,ando2016insect}. We call such controllers wind followers, because they
treat the wind sensed now as a pointer back to the source.

\begin{figure}[t]
\centering
\includegraphics[width=\textwidth]{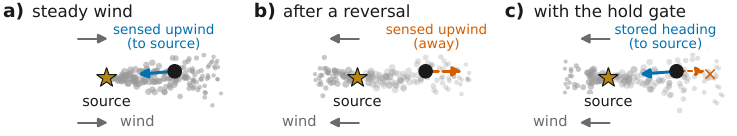}
\caption{\textbf{A hold gate steers by the wind stored at contact when the live wind contradicts it.}
\textbf{a)} In steady wind odour lies downwind, so surging up the sensed wind (solid blue arrow)
approaches the source. \textbf{b)} After the wind reverses, odour released earlier still surrounds the
robot while the sensed upwind points away (dashed vermilion arrow). \textbf{c)} The hold gate rejects a
live wind that contradicts the heading stored when contact began and passes the follower the stored
heading. Schematic; filaments are illustrative.}
\label{fig:concept}
\end{figure}

That pointer is right only while the air around the robot still moves along the path the odour
took. When the wind turns, odour released before the turn keeps arriving although the sensed wind
already points elsewhere (Fig.~\ref{fig:concept}a,b). Such turns matter outdoors, where moths reach
pheromone sources even though the puffs they follow have drifted away from the mean
wind~\cite{david1983finding,elkinton1987pheromone}. Indoors, a fan or ventilation system that
switches could plausibly turn the air in the same way, although we know of no measurement of how
often indoor airflow reverses. The obvious remedy,
steering by a heading taken at an earlier contact, fails in a different way, because a heading taken
from old odour points the wrong way for as long as it is kept and carries the robot away from the
plume between contacts.

Earlier work has treated the sensed wind with more care than a single reading and has kept other
memories. Robots have estimated the wind from several anemometric sensors~\cite{ishida1994study} or
on a micro-drone~\cite{neumann2012autonomous} and modelled how the flow varies in space and
time~\cite{hernandezbennetts2017probabilistic}. A moth-inspired tracer on an underwater vehicle
steered by a flow estimate averaged over \SI{100}{\second} and searched around its last detection %
once contact had lapsed~\cite{li2006moth}. Probabilistic searchers go further and infer the source
from detections and the history of the measured flow, using source-likelihood
maps~\cite{farrell2003plume,pang2006chemical}, particle filters~\cite{li2011odor,neumann2013gas} or
information-seeking
strategies~\cite{vergassola2007infotaxis,masson2013olfactory,ristic2016study,hutchinson2018entrotaxis,bourne2019coordinated},
although most of them need the robot's position. In the work we reviewed
(refs.~\cite{li2006moth,farrell2003plume,pang2006chemical,li2011odor,neumann2013gas,vergassola2007infotaxis,bourne2019coordinated}),
none substitutes the flow direction recorded at contact for the live one when the two disagree.

We show below that following the live wind and holding an earlier heading lose ground in different
sensing states, so a controller need not choose between them. A hold gate keeps the sensed upwind from the start of contact and, only while the live
wind contradicts it by more than a right angle, passes the stored heading to the unchanged follower,
which we call its host (Fig.~\ref{fig:concept}c). The gate needs no map, odometry or source
position. It should help wherever the stored heading points to the source more often than the live
wind does at the moments the gate acts. We test this rule in simulation, using trials on a legged robot only
to confirm that a reversal turns the robot's sensed wind against the source, not to run On reversing flows drawn at random the gate raised every follower we tested. At the moth model's own
scale, however, a plain delay of the wind input was not resolvably worse, and the gate beat the best
delay only in a rescaled simulation with the robot's sensors and turn-rate limit, by a margin that
survives correction over all comparisons for one of the two followers tested.
Since a plain spiral searcher still outperformed the gated moth model, the gate improves a follower
rather than replacing a better searcher.

\section{Results}

\subsection{Why reversals mislead wind followers}

\begin{figure}[t]
\centering
\includegraphics[width=0.85\textwidth]{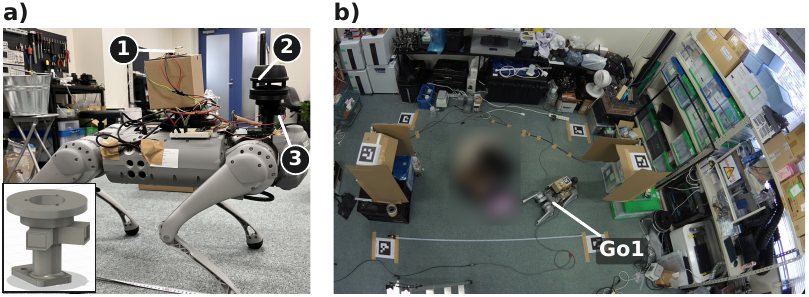}
\caption{\textbf{The rig used to check that a reversal turns the sensed wind against the source.}
\textbf{a)} The Go1: 1, the two-dimensional thermal anemometer; 2, an unused ultrasonic anemometer;
3, a metal-oxide gas sensor, one on each side; inset, the gas-sensor mast. \textbf{b)} The arena from
the overhead scoring camera, a fan bank at each end (operator blurred). Runs and outcomes are in
Supplementary Fig.~S1 and the Supplementary Information.}
\label{fig:robot}
\end{figure}

\begin{figure}[t]
\centering
\includegraphics[width=0.82\textwidth]{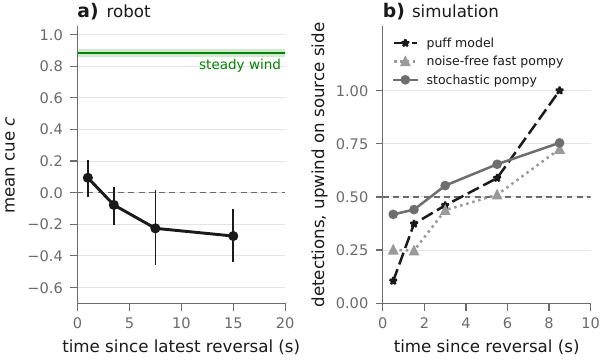}
\caption{\textbf{After a reversal, the sensed wind comes to point away from the source.}
\textbf{a)} Robot: cosine between sensed upwind and the bearing to the source against time since the
latest reversal; mean over the fifteen reversing runs in bins of 0 to 2, 2 to 5, 5 to 10 and 10 to
20~s, plotted at bin midpoints, with 95\% bootstrap intervals over runs; band, steady wind.
\textbf{b)} Simulation without a controller: fraction of detections with upwind on the source's side
against time since reversal, for a stochastic pompy field, the same field without boundary noise
(both $k=320$) and a puff model that moves the whole plume with the wind.}
\label{fig:cue}
\end{figure}

We first checked that the failure occurs on a real sensor. A Unitree Go1 quadruped walked between
two fan banks that swapped every \SI{20}{\second}, from a start near one bank towards an ethanol %
source near the other, so whenever the start-side bank blew, the true upwind at the robot pointed
away from the source (Fig.~\ref{fig:robot}; Methods). We recorded the cue $c$, the cosine between
the upwind reported by the robot's anemometer and the camera bearing to the source. In 15 %
steady-wind runs $c$ averaged \num{0.88} (95\% CI \numrange{0.86}{0.91}), but in 15 reversing %
runs it averaged \num{-0.27} (\numrange{-0.44}{-0.11}) \SIrange{10}{20}{\second} after a reversal %
(Fig.~\ref{fig:cue}a; Supplementary Table~S1). In the 29 of 40 reversals that settled before the %
next, the sensed wind took a median of \SI{10.7}{\second} to settle, far longer than the %
anemometer's \SI{0.51}{\second} reading interval, which is consistent with the arena's air turning %
rather than with sensor lag (Supplementary Table~S2).
These trials are an observation rather than a test of strategies, because each controller ran only
three times from one start, the gas reading exceeded threshold on about \SI{99}{\percent} of ticks, %
and the score was adopted after the trials (Supplementary Table~S3; Supplementary Fig.~S1).

For a stationary source, geometry explains why a reversal turns the sensed wind against the
source. Sensed upwind points to the source's side
exactly when the robot stands downwind of the source in the wind it senses now,
\begin{equation}
\mathbf{u}(t)\cdot(\mathbf{x}-\mathbf{s}) > 0 ,
\label{eq:criterion}
\end{equation}
where $\mathbf{u}(t)$ is the wind at the robot's position $\mathbf{x}$ and $\mathbf{s}$ is the
source. In steady wind odour exists only downwind, so every detection satisfies
Eq.~\eqref{eq:criterion}. After a reversal, however, odour released under the old wind lies on the
side from which the new wind blows, and every detection of it points away from the source. How much
of that odour remains depends on the flow, which we measured with a stationary probe in simulation
that recorded, at every detection near the source, whether the local upwind pointed to the source's
side (Methods). In pompy, a filament-based plume model with a stochastic wind
field~\cite{farrell2002filament}, the local upwind still pointed to the source's side at %
\numrange{0.42}{0.74} of detections in the first second after a reversal, across five values of the %
field diffusivity $k$, which sets how quickly the field follows a reversal. In the same first second
the share was lower, \num{0.25}, in the field without boundary noise at $k=320$, and lower still, %
\num{0.11}, in a puff model that moves the whole plume with the wind (Fig.~\ref{fig:cue}b), since a smooth flow carries the old plume back
over the source intact. The same signature appears in a published outdoor
dataset~\cite{nag2024odour}, in which the mean cue at odour detections turned from \num{0.45} to %
\num{-0.49} (\numrange{-0.65}{-0.28}) within \SIrange{2}{5}{\second} of a forest wind shift, %
although the interval reaches zero when the nearest anemometer is used instead of the source's
(Supplementary Table~S4).

\begin{figure}[t]
\centering
\includegraphics[width=\textwidth]{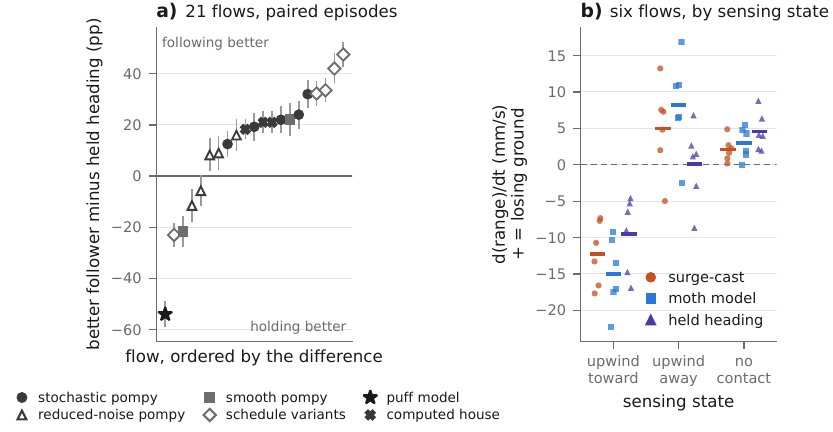}
\caption{\textbf{Following and holding each win in some flows and lose ground in different sensing
states.} \textbf{a)} For each of 21 reversing flows, the better of two wind followers minus the held %
heading on the same 400 episodes (paired difference) with its 95\% paired bootstrap interval over
episodes; flows ordered by the difference. Stochastic pompy, boundary-noise gain 2; reduced-noise,
0.5 or 1; smooth, 0. \textbf{b)} Rate of change of range to the source by sensing state (contact with
sensed upwind toward or away from the source, or no contact) for three controllers on the six
diagnosis flows (dots, one per flow; lines, mean over flows), on 40 selection episodes per flow held %
out from every test; positive is losing ground.}
\label{fig:fail}
\end{figure}

This geometry implies that following the live wind and holding an earlier heading fail in different
places. To check, we compared a held heading with two wind followers, surge-cast and a published
model of silkmoth search~\cite{yamada2021multisensory}, which we call the moth model. The held
heading steers by the upwind stored at contact and replaces it at the next contact after a gap
(Methods). The benchmark works at
the moth model's own scale, with an insect-scale simulated quadruped that surges at %
\SI{22}{\milli\metre\per\second}, starts \SI{300}{\milli\metre} from a source it cannot see in a %
\SI{160}{\milli\metre\per\second} wind, and succeeds if it comes within \SI{50}{\milli\metre} of it. %
Across 21 reversing flows the better follower beat the held heading in most, but holding won, %
mostly where weak mixing let the old plume survive a reversal (Fig.~\ref{fig:fail}a; Supplementary Table~S5). In
the puff model, for example, the held heading reached the source in \SI{97.0}{\percent} of episodes %
and the moth model in only \SI{43.0}{\percent}. The reason appears in tick-by-tick logs on six %
diagnosis flows, which show that during contact whose sensed wind pointed away from the source the
moth model lost ground at
\SI{8.2}{\milli\metre\per\second} while the held heading barely moved %
(\SI{0.1}{\milli\metre\per\second}). Out of contact, by contrast, the held heading lost ground %
faster, at \SI{4.6}{\milli\metre\per\second} against the moth model's %
\SI{3.0}{\milli\metre\per\second}, so holding protects where following fails but costs where %
following is safe (Fig.~\ref{fig:fail}b).

The hold gate keeps only the part of the held heading that protects (Algorithm~\ref{alg:gate}). At
the first detection it stores the sensed upwind as a world-frame heading, counting a detection only
while the robot heads within \SI{120}{\degree} of the sensed upwind, and it discards that heading %
after \SI{2}{\second} without detection. While a heading is stored and the live upwind differs from %
it by more than \SI{90}{\degree}, the host receives the stored heading in place of the live wind. %
In every other case, including after the memory is discarded, the host runs unchanged. All three
constants come from the held heading, and the host and the form of the gate were chosen on
selection episodes disjoint from every test (Methods).

\subsection{The gate raised every follower on random flows, but at insect scale a delay was not resolvably worse}

We tested the gate on 60 pompy flows drawn at random, with reversal separations between %
\SI{120}{\degree} and \SI{180}{\degree}, square-wave or Poisson-timed reversals at a mean interval %
of \SIrange{2.5}{20}{\second}, release ranges of \SIrange{300}{600}{\milli\metre}, and random field %
irregularity and diffusivity (Methods). Every comparison is paired, since the controllers ran on the
same episodes, and an effect is the mean over flows of the per-flow difference in success, in
percentage points (pp), with its 95\% interval in brackets from a crossed bootstrap over flows and
episodes (Methods). On 150 paired episodes per flow, %
\SI{50.63}{\percent} to \SI{61.40}{\percent}, by \num{10.77}~pp \ci{6.60}{15.39} (Holm-adjusted %
$P\le\num{1.5e-4}$; Table~\ref{tab:random}). Separate tests on the same flows, each on its own %
episodes, raised the four other hosts by \numrange{5.97}{10.13}~pp, with every interval above zero %
(Fig.~\ref{fig:mech}a; Table~\ref{tab:random}). Among these four, the largest gains went to a
count-gated surge and to a silkmoth surge-zigzag-loop model (SZL)~\cite{kanzaki1992zigzag}, the
smallest to surge-cast
and to a surge with an expanding spiral search, which we call the spiral host. These gains do not
reflect a choice of favourable hosts, because no host was tuned to these flows (Supplementary
Table~S6) and small screens on chosen flows had even shown the gate lowering the spiral host, SZL
and surge-cast (Methods). The two largest gains, however, went to weak searchers, since the
count-gated surge and SZL reached the source in fewer than a quarter of episodes even when gated. %

The choice of host mattered more than the gate, since the plain spiral host reached the source in %
\SI{74.87}{\percent} of episodes and beat the gated moth model by \num{15.37}~pp %
\ci{6.57}{24.67} (Table~\ref{tab:random}). Because this comparison was planned as a test for a %
gain of the gated moth model, we report it two-sided and post hoc. Controllers that use the history
of the wind in other ways did worse, even when tuned on a separate screen of the random-flow
distribution. One steered by the sensed upwind
averaged over a window and the other toward the peak of a source-likelihood map traced back from
detections~\cite{pang2006chemical}, to which we added a map decay. Tuning (Supplementary Table~S7)
raised the averaging controller's success from \SI{7.6}{\percent} to \SI{16.8}{\percent} and the %
likelihood controller's from \SI{25.3}{\percent} to \SI{42.5}{\percent}. The gated moth model %
still beat them by \num{41.17} and \num{15.50}~pp (Table~\ref{tab:random}), although the tuned %
likelihood horizon lay at the edge of its grid.

A follower that receives the wind late already holds part of the old wind, which raises the question
of whether simply smoothing or delaying the wind would do the gate's work. A first-order low-pass of
the moth model's wind input, with its time constant (\SI{1}{\second}) chosen on a separate block, %
did not resolvably change the moth model's success ($+0.30$~pp \ci{-3.40}{4.10}, descriptive), and %
the gated moth model beat the low-passed one by \num{9.07}~pp (Holm $P=\num{8.0e-4}$). In the %
later delay test the gate's margin over a re-chosen low-pass was smaller, \num{5.5}~pp with the %
benchmark's anemometer (Holm $P=\num{0.069}$) and \num{4.0}~pp with the Go1's (Holm %
$P=\num{0.049}$), and neither passes the Bonferroni check. A delay is a closer rival, because it passes the follower %
the wind sensed a fixed time earlier and is therefore, like the gate, a memory of the earlier wind.
A screen of five delays (Supplementary Table~S8) chose \SI{2}{\second} with the benchmark's %
anemometer and \SI{0.25}{\second} with a model of the Go1's noisier anemometer and gait sway at %
insect scale. With the benchmark's anemometer the delay itself raised the moth model by %
\num{5.4}~pp (descriptive), and the gate's margin over it, \num{1.61}~pp \ci{-3.28}{6.72} (Holm %
$P=\num{0.52}$), resolved no difference. With the Go1's anemometer the margin was larger, %
\num{2.61}~pp, but still not resolved (Holm $P=\num{0.090}$). At insect scale, then, a cruder %
memory of the earlier wind recovered most of the gate's gain. Combining the two did not help
either, since gating the delayed follower changed its success by $-7.17$~pp \ci{-11.56}{-2.94} %
with the benchmark's anemometer (Table~\ref{tab:random}).

\begin{table}[t]
\caption{\textbf{On the 60 random insect-scale flows the gate raised every host and beat the tuned
wind-history controllers, but it beat the low-pass less firmly, did not resolvably beat a delay, and
lost to the plain spiral host.} The table lists every main comparison on these flows. Each row is a paired comparison on its own block of
episodes, so success levels pair only within a row. Difference: first minus second controller, mean
over flows of the per-flow paired difference, with a 95\% interval from a crossed bootstrap that
resamples flows and episodes independently (Methods). $P$: one-sided bootstrap $P$ for a gain of the
first controller, Holm-adjusted within its test; $^{\ddagger}$, two-sided, post hoc, for an estimate
below zero. Delays and time constants were chosen on separate screening blocks. Per-test tables:
Supplementary Table~S9 (moth model, surge-cast), Supplementary Table~S10 (count-gated surge, SZL, per flow), Supplementary Table~S11
(surge-cast host), Supplementary Table~S12 and Supplementary Table~S13 (low-pass, spiral host), Supplementary Table~S14 (tuned
controllers) and Supplementary Table~S15 (delay).}
\label{tab:random}
{\footnotesize\setlength{\tabcolsep}{3pt}
\begin{tabular}{@{}lrrrr@{}}
\toprule
Comparison & Success (\%) & Episodes & Difference (pp) & $P$ (Holm) \\
 & first / second & per flow & [95\% interval] & \\
\midrule
\multicolumn{5}{@{}l}{\emph{Gated minus plain host}} \\
moth model & 61.4 / 50.6 & 150 & +10.8 [+6.6, +15.4] & $\le\!1.5\!\times\!10^{-4}$ \\
count-gated surge & 19.1 / 8.9 & 50 & +10.1 [+6.7, +14.0] & $\le\!1.5\!\times\!10^{-4}$ \\
SZL & 23.2 / 13.2 & 50 & +10.0 [+6.5, +13.9] & $\le\!1.5\!\times\!10^{-4}$ \\
surge-cast & 43.0 / 34.9 & 50 & +8.1 [+3.4, +12.9] & $5.0\!\times\!10^{-4}$ \\
spiral host & 80.8 / 74.9 & 50 & +6.0 [+2.0, +10.1] & $6.3\!\times\!10^{-3}$ \\
\midrule
\multicolumn{5}{@{}l}{\emph{Gated moth model minus another controller}} \\
surge-cast & 61.4 / 31.9 & 150 & +29.5 [+25.6, +33.5] & $\le\!1.5\!\times\!10^{-4}$ \\
averaging, tuned & 58.0 / 16.8 & 30 & +41.2 [+33.1, +49.2] & $\le\!1.0\!\times\!10^{-4}$ \\
source-likelihood, tuned & 58.0 / 42.5 & 30 & +15.5 [+9.2, +22.3] & $\le\!1.0\!\times\!10^{-4}$ \\
low-passed moth model ($\tau=1$~s) & 59.5 / 50.4 & 50 & +9.1 [+3.9, +14.6] & $8.0\!\times\!10^{-4}$ \\
plain spiral host & 59.5 / 74.9 & 50 & $-$15.4 [$-$24.7, $-$6.6] & 1.00; $5.0\!\times\!10^{-4}$$^{\ddagger}$ \\
low-passed gated $-$ low-passed & 53.3 / 50.4 & 50 & +2.8 [+0.8, +5.0] & $6.3\!\times\!10^{-3}$ \\
\midrule
\multicolumn{5}{@{}l}{\emph{Against the best delay and low-pass: benchmark anemometer}} \\
gated $-$ delayed ($L=2$~s) & 48.8 / 47.2 & 30 & +1.6 [$-$3.3, +6.7] & 0.52 \\
gated $-$ low-passed ($\tau=1$~s) & 48.8 / 43.3 & 30 & +5.5 [+0.2, +11.2] & 0.069 \\
gated on delayed $-$ delayed & 40.1 / 47.2 & 30 & $-$7.2 [$-$11.6, $-$2.9] & 1.00 \\
\midrule
\multicolumn{5}{@{}l}{\emph{The same with the Go1's anemometer and gait sway}} \\
gated $-$ delayed ($L=0.25$~s) & 52.4 / 49.8 & 30 & +2.6 [$-$0.8, +6.1] & 0.090 \\
gated $-$ low-passed ($\tau=0.5$~s) & 52.4 / 48.4 & 30 & +4.0 [+0.3, +7.8] & 0.049 \\
gated on delayed $-$ delayed & 52.8 / 49.8 & 30 & +3.0 [$-$0.4, +6.6] & 0.090 \\
\bottomrule
\end{tabular}}

\end{table}

The gate's cost in steady wind, where there is no old odour to correct, was small but not zero. On
100 paired steady-wind episodes per host at insect scale it lowered the moth model's success from %
\SI{98}{\percent} to \SI{93}{\percent}, a difference of $-5$~pp \ci{-10}{-1}, and lowered the other %
four hosts by \numrange{1}{5}~pp (Supplementary Table~S16; descriptive).

An earlier test on 31 flows that we chose to span the regimes, with 400 paired episodes per flow, %
agreed with the random-flow test in raising the moth model, by \num{8.73}~pp \ci{1.25}{16.61} %
(Holm $P=\num{0.020}$; Supplementary Table~S17). The gated moth model's margin over surge-cast, %
however, which had excluded zero when only episodes were resampled, no longer did once flows were
resampled too (\num{3.44}~pp \ci{-3.18}{11.49}). That margin came from the two puff-model flows %
(Supplementary Table~S18).

\subsection{The gate helps where the stored heading beats the live wind and the host still fails}

\begin{figure}[!tp]
\centering
\includegraphics[width=\textwidth]{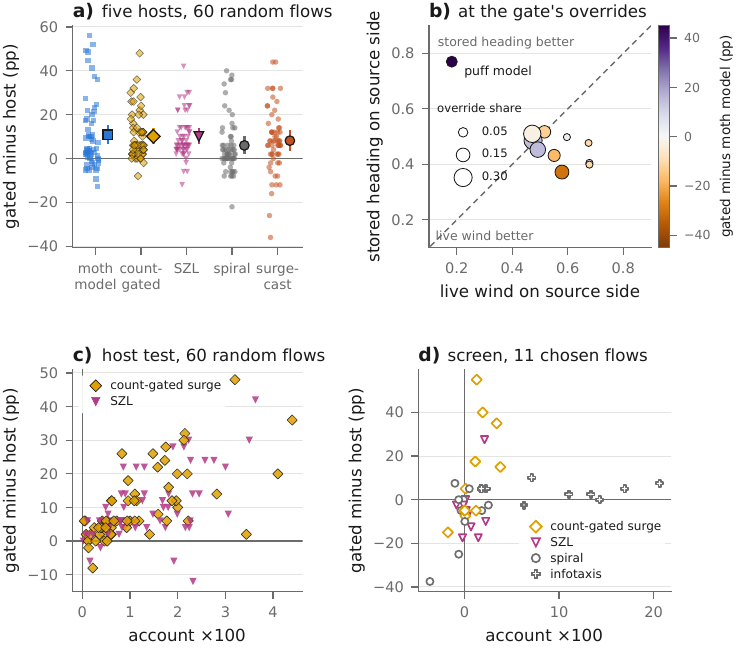}
\caption{\textbf{The gate raises five hosts on average, and its gain correlates with the stored
heading's advantage at its overrides.} \textbf{a)} Gated minus plain host on the 60 random flows,
one small mark per flow; large marks, mean over flows with its 95\% crossed-bootstrap interval (moth %
model 150, other hosts 50 paired episodes per flow, each host on its own block). \textbf{b)} At the
gate's overrides of the moth model, the share of override ticks on which the stored heading and the
live wind pointed to the source's side, one point per chosen flow (eleven; the three computed-house
variants sit near a live-wind share of 0.68), on selection episodes; size, override share; colour, %
gated minus plain moth model. \textbf{c)} Gated minus plain host against the gate's account (override
share times the stored heading's advantage), one point per flow and host, for the count-gated surge
and SZL on the 60 random flows (120 cells). \textbf{d)} The same for four hosts on an exploratory %
screen of eleven chosen flows (44 cells); a different population from c, not pooled with it.} %
\label{fig:mech}
\end{figure}

Because the gate replaces one direction with another, its effect should depend on how often each
of them points to the source when it acts. We summarise this in the gate's account of a flow, the
override share (the share of ticks on which the host sees the stored heading) multiplied by the
stored heading's advantage (the share of overrides on which the stored heading lay on the source's
side minus the share on which the live wind did; Methods). Since the account needs the source's
position, it is a diagnostic rather than a signal a robot could use. On selection episodes of eleven
chosen flows the gate overrode mostly while its host surged. Where the stored heading had the %
advantage it could be large, as in the puff model, where it lay on the source's side at \num{0.77} %
of overrides against \num{0.18} for the live wind (Fig.~\ref{fig:mech}b; Supplementary %
Table~S19). On two flows where the stored heading was right less often than the live wind, the gate
lowered its host, by \num{38.75}~pp when reversals came every %
\SI{2.5}{\second} (stored \num{0.37} against live \num{0.58}) and by \num{24.00}~pp with %
Poisson-timed reversals in a noise-free field (\num{0.43} against \num{0.55}). The relation is not strict, however, since on a noise-free flow %
with $k=320$ the gate raised the moth model by \num{12.50}~pp although the stored heading was right %
less often (\num{0.45} against \num{0.49}). %

Across flows and hosts the gain rank-correlated with the account, with Spearman %
$\rho=\num{0.614}$ over the 120 flow-by-host cells of the count-gated surge and SZL %
($P\le\num{5e-5}$ with cells permuted, and also with whole flows permuted, post hoc;
Fig.~\ref{fig:mech}c). Because account %
and gain there came from the same episodes, we repeated the analysis for surge-cast and the moth
model with each cell's episodes split by index parity, relating the account on one half to the gain
on the other, and still found $\rho=\num{0.307}$ ($P=\num{4.0e-4}$ with flows permuted, post hoc; %
Supplementary Tables~S11 and~S20). The account is close to definitional, since the gate can gain
only by overriding when the stored heading is right. The cross-fitted correlation therefore confirms
this reading on episodes from which the account was not computed, but it does not demonstrate a
cause. The account is also an imperfect predictor, having overstated the gain for the spiral host and infotaxis on a screen of eleven chosen
flows (Fig.~\ref{fig:mech}d).

\begin{figure}[!tp]
\centering
\includegraphics[width=\textwidth]{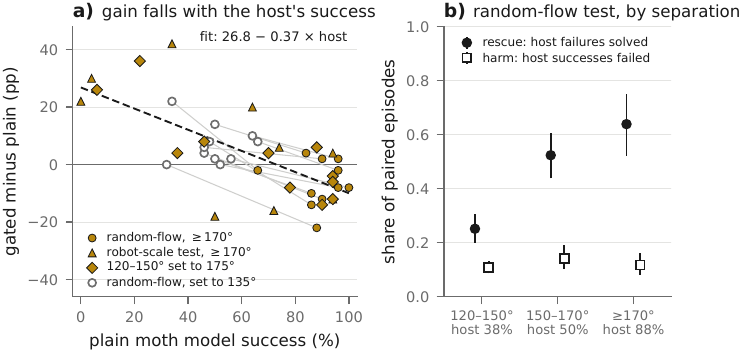}
\caption{\textbf{At near-full reversals the gate gained little because the plain moth model rarely
failed, not because it rescued less.} \textbf{a)} Gated minus plain moth model against the plain model's
success, one point per flow and separation of an exploratory diagnosis at insect scale, 50 paired %
episodes each: filled, flows reversing by about \SI{175}{\degree}; open, the random flows of
\SI{170}{\degree} or more set to \SI{135}{\degree}, each joined to its own flow at its own
separation; dashed, least-squares line over all 42 points. \textbf{b)} On the random-flow test's %
paired episodes, by reversal separation: rescue, the share of the host's failed episodes that the
gated form solved; harm, the share of its successful episodes that the gated form failed; 95\%
intervals resampling flows then episodes. Below each bin, the plain moth model's success.}
\label{fig:nearfull}
\end{figure}

Split by schedule, most of the gain came from square-wave reversals, where the gate raised the moth
model by \num{20.96}~pp \ci{13.25}{28.77} on the 25 square-wave flows but by %
only \num{3.49}~pp \ci{0.80}{6.30} on the 35 Poisson-timed ones, and the other hosts showed the %
same pattern (descriptive and post hoc, with intervals that resample flows and then episodes;
Supplementary Table~S21). At reversals of \SI{170}{\degree} or more (post hoc, 11 flows) the gate gave the moth %
model nothing ($-2.91$~pp \ci{-6.73}{1.09}), and there the account did not track the gain %
($\rho=\num{0.10}$ over 22 cells). An exploratory diagnosis with every tick logged traced this to %
headroom (Fig.~\ref{fig:nearfull}; Supplementary Table~S22), because on those flows the plain moth
model already reached the source in \SI{88.2}{\percent} of episodes.

When we set the same flows to a %
\SI{135}{\degree} separation with all else fixed, the host's success fell by \num{39.5}~pp and %
the gate's gain rose by \num{13.3}~pp \ci{9.6}{17.3}. Across 42 insect-scale cells the gain fell by %
\num{0.37}~pp \ci{0.26}{0.47} for every point of host success, and the near-full cells lay on that %
line. Measured against the host's failures, the gate in fact did more at near-full reversals than
elsewhere, turning \num{0.64} \ci{0.52}{0.75} of them into successes against \num{0.25} at %
\SIrange{120}{150}{\degree}. Its harm, by contrast, did not depend on separation, since it turned
\numrange{0.11}{0.14} of the host's successes into failures at every separation. Because the host %
succeeded nine times in ten at near-full reversals, that small harm outweighed the rescues. The %
logs suggest that the harm arises in near-full flips, where old odour washes back
along the same axis, contact continues, and the gate keeps overriding with a stale heading
(override share \num{0.095} against \num{0.020} at \SI{135}{\degree}). %

The gate also failed where the stored heading had no advantage. On the 31 chosen flows it lowered %
its own host in 8 with a paired interval below zero. Seven of these survived Holm-corrected McNemar %
tests, and the exception was the default flow (pompy noise 2, $k=20$), with a difference of %
\num{-6.75}~pp \ci{-11.26}{-2.25} and $P=\num{0.067}$ (Supplementary Table~S23). The account was %
logged for six of the eight, and in each of them the stored heading was right about as often as or %
less often than the live wind,
whereas for the other two (reversals every \SI{20}{\second}, and Poisson-timed reversals in the %
default field) we did not establish why. In a computed indoor flow, a house switching between two
ventilation solutions, the gate was nearly idle and the live wind was the better guide. The gate
nonetheless lowered the moth model's success when the house was solved through the switch with the
flow's own transient response, by \num{8.0}, \num{3.5} and \num{2.5}~pp at switching periods of 10, %
20 and \SI{40}{\second} (exploratory; exact McNemar $P=\num{0.0025}$, \num{0.12} and \num{0.13}; %
Supplementary Table~S24). Finally, a model of the Go1's anemometer removed the gain on
the six diagnosis flows ($-0.8$~pp \ci{-6.7}{5.0}, exploratory; Supplementary Table~S25), but those %
flows had been chosen to show where the gate loses. On 60 new random insect-scale flows with the %
same anemometer and gait sway, the gate raised the moth model from \SI{43.9}{\percent} to %
\SI{58.3}{\percent}, by \num{14.47}~pp \ci{9.80}{19.27} (Table~\ref{tab:robotscale}), so the %
earlier null belonged to the flows rather than to the sensor.

\subsection{In a rescaled simulation with the robot's sensors and turn rate, the gate beat the best delay}

A simulation of the fan arena itself could not test the gate, because with saturated gas old and new
odour are indistinguishable, and with the plume's own gas the followers met almost no odour after
reversals (Supplementary Tables~S26 and~S27). We therefore rescaled the insect-scale benchmark to
the robot's size, multiplying every length by \num{7.5} and leaving time unchanged. Geometry %
and flow timing then match the insect-scale benchmark by construction, and the benchmark's wind
becomes \SI{1.2}{\metre\per\second}, the arena wind the robot's anemometer read at its start. As a %
check, with the benchmark's own sensors the rescaled simulation reproduced insect-scale episodes
exactly (Methods; Supplementary Table~S28). The simulated robot then carried the Go1's anemometer model, heading
drift and gas-sensor placement, and walls stopped it. This remains a planar plume model rather than
a robot-scale environment, and it differs from the robot in three ways that matter for the gate: its
gas sensors respond instantly, its wind turns within seconds of a switch whereas the arena's air took
longer, and its plume spread rests on an assumed turbulence intensity~\cite{taylor1922diffusion}. We
added the robot's properties in steps, each a new test on new flows drawn from the random-flow
distribution (Table~\ref{tab:robotscale}).

\begin{table}[t]
\caption{\textbf{In the rescaled simulation the gate raised its hosts in every setting, and with the
robot's turn cap it beat the best delay.} The table lists every main comparison of the robot-like tests. Each
block of rows is a test on its own 60 flows; 25 paired episodes per flow and controller; the last %
row is at insect scale. Difference: first minus second controller, mean over flows of the per-flow
paired difference, with a 95\% crossed-bootstrap interval (Methods). $P$: bootstrap $P$, one-sided
for a gain unless marked $^{\dagger}$ (two-sided), Holm-adjusted within its test. Delays were chosen
per host on a separate screen. Per-test tables: Supplementary Table~S29, Supplementary Table~S30, Supplementary Table~S31 and Supplementary Table~S32.}
\label{tab:robotscale}
{\footnotesize\setlength{\tabcolsep}{3pt}
\begin{tabular}{@{}lrrrr@{}}
\toprule
Comparison & Success (\%) & Episodes & Difference (pp) & $P$ (Holm) \\
 & first / second & per flow & [95\% interval] & \\
\midrule
\multicolumn{5}{@{}l}{\emph{Gated minus plain host: Go1 sensors, assumed sway, the controllers' turn rate}} \\
moth model & 48.9 / 30.5 & 25 & +18.3 [+12.5, +24.5] & $\le\!1.5\!\times\!10^{-4}$ \\
surge-cast & 24.8 / 9.3 & 25 & +15.5 [+10.3, +21.1] & $\le\!1.5\!\times\!10^{-4}$ \\
spiral host & 70.8 / 57.3 & 25 & +13.5 [+8.6, +18.7] & $\le\!1.5\!\times\!10^{-4}$ \\
\midrule
\multicolumn{5}{@{}l}{\emph{Gated minus plain host: + the Go1's turn cap and estimated sway}} \\
moth model & 51.1 / 43.5 & 25 & +7.6 [+3.1, +12.3] & $1.7\!\times\!10^{-3}$ \\
spiral host & 65.5 / 58.6 & 25 & +6.9 [+2.0, +11.9] & $3.1\!\times\!10^{-3}$ \\
\midrule
\multicolumn{5}{@{}l}{\emph{Gated minus plain host: + operating-wind anemometer noise}} \\
moth model & 60.7 / 48.8 & 25 & +11.9 [+7.0, +17.0] & $\le\!1.0\!\times\!10^{-4}$ \\
spiral host & 70.3 / 60.3 & 25 & +10.0 [+5.3, +14.7] & $\le\!1.0\!\times\!10^{-4}$ \\
\midrule
\multicolumn{5}{@{}l}{\emph{Same setting: against the best delay}} \\
moth model: gated $-$ delayed ($L=2$~s) & 56.3 / 50.5 & 25 & +5.8 [+0.9, +10.9] & 0.024$^{\dagger}$ \\
spiral host: gated $-$ delayed ($L=0.5$~s) & 69.8 / 60.7 & 25 & +9.1 [+4.5, +13.9] & $2.4\!\times\!10^{-3}$$^{\dagger}$ \\
moth model: gated on delayed $-$ delayed & 44.1 / 50.5 & 25 & $-$6.4 [$-$11.5, $-$1.5] & 0.024$^{\dagger}$ \\
spiral host: gated on delayed $-$ delayed & 67.4 / 60.7 & 25 & +6.7 [+1.9, +11.7] & 0.024$^{\dagger}$ \\
\midrule
\multicolumn{5}{@{}l}{\emph{Insect scale, random flows: Go1 anemometer and sway}} \\
moth model, gated minus plain & 58.3 / 43.9 & 25 & +14.5 [+9.8, +19.3] & $\le\!5.0\!\times\!10^{-5}$ \\
\bottomrule
\end{tabular}}

\end{table}

With the Go1's sensors, an assumed gait sway and the controllers' own turn-rate limit, the gate
raised the moth model by \num{18.33}~pp \ci{12.53}{24.53} and raised surge-cast and the spiral %
host by \num{15.47} and \num{13.53}~pp (Table~\ref{tab:robotscale}). The cross-fitted account %
again correlated with the gain ($\rho=\num{0.193}$, $P=\num{5.0e-4}$). The controllers' turn rate %
of \SI{8}{\radian\per\second}, however, is sixteen times the \SI{0.5}{\radian\per\second} at which %
the robot's control software clipped yaw commands, so the next test clipped every commanded turn at
that rate. The same test raised the gait sway to \SI{410}{\milli\metre\per\second}, a value estimated from %
the robot's logs that errs high if its anemometer follows the gait more closely than our lag model
(Methods). The gate still raised the moth model by \num{7.60}~pp \ci{3.07}{12.27} (Holm %
$P=\num{0.0017}$) and the spiral host by \num{6.87}~pp (Holm $P=\num{0.0031}$). Surge-cast was %
left out of this test, because in a screen at the capped turn rate it reached the source in only
\SI{2}{\percent} of episodes plain and \SI{12}{\percent} gated. Its \SI{1.2}{\radian} cast swings %
take about \SI{5}{\second} at that rate against a \SI{2}{\second} cast half-period, so a follower %
whose casting needs fast turns cannot search at the Go1's turn rate, gated or not. When the
anemometer noise was lowered from its near-still-air value, \SI{21.3}{\degree}, to its standing %
value in the operating wind, \SI{6.8}{\degree}, the gate raised the moth model by \num{11.93}~pp %
\ci{7.00}{17.00} and the spiral host by \num{10.00}~pp (Holm $P\le\num{1.0e-4}$ each). In 100 %
paired steady-wind episodes per host at this setting it changed no outcome, which bounds its
steady-wind cost there at \num{3.6}~pp (exact 95\% bound, post hoc). %

\begin{figure}[!tp]
\centering
\includegraphics[width=\textwidth]{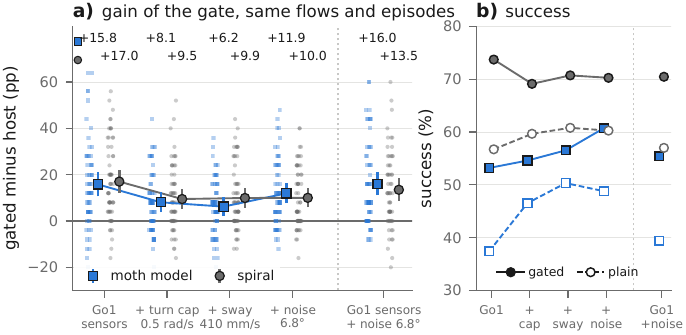}
\caption{\textbf{On the same flows and episodes, the gate raised both hosts in every robot-scale
setting.} The moth model and the spiral host, plain and gated, on 60 flows with 25 paired episodes %
per flow, in five settings of the rescaled simulation: the Go1's sensors with an assumed gait sway %
and the controllers' own turn rate (Go1); with the Go1's turn cap added (+cap); with the gait sway
raised to the value estimated from its logs (+sway); with the anemometer noise lowered to its
operating-wind value (+noise); and, right of the dashed line, the first setting with only the noise
lowered. \textbf{a)} Gated minus plain host: small marks, one per flow; large marks and the labels
above (moth model upper, spiral lower), mean over flows with a descriptive 95\% interval resampling
flows then episodes. \textbf{b)} Success of the plain (open) and gated (filled) host, averaged over
flows.}
\label{fig:robotscale}
\end{figure}

Because these tests ran on different flows, their gains cannot be compared directly, so a paired
comparison reran the moth model and the spiral host on one set of 60 flows and episodes with the
robot's properties added one at a time (Fig.~\ref{fig:robotscale}; Supplementary Table~S33). The
gate raised both hosts in every setting, by \numrange{6.2}{17.0}~pp. Adding the turn cap changed %
the gain by $-7.73$~pp \ci{-14.53}{-1.20} for the moth model and by a similar $-7.53$~pp for the %
spiral host. Although both intervals lay below zero, neither change survived Holm adjustment over a
host's four steps (two-sided $P=\num{0.088}$ and \num{0.10}), so the cap is a likely rather than %
an established cause of the lower gains. For the moth model the cap acted mostly by letting the
plain host search better, raising its success from \SI{37.4}{\percent} to \SI{46.5}{\percent}. %
Neither the estimated sway nor the operating-wind noise changed either gain resolvably after Holm
adjustment.

With the turn cap, estimated sway and operating-wind noise all in place, we finally compared the gate
with the best fixed delay, which a screen chose per host from five candidates as \SI{2}{\second} %
for the moth model and \SI{0.5}{\second} for the spiral host (Supplementary Table~S34). On 60 new %
flows the gate beat the delay by \num{5.80}~pp \ci{0.87}{10.93} for the moth model (two-sided %
Holm $P=\num{0.024}$ over the test's four comparisons) and by \num{9.13}~pp for the spiral host %
(Holm $P=\num{0.0024}$; Table~\ref{tab:robotscale}). Only the spiral host's margin, however, also %
passes the Bonferroni check over all of the paper's comparisons (Methods). Gating the delayed %
follower changed the moth model by $-6.40$~pp \ci{-11.47}{-1.47}, as with the benchmark's %
anemometer at insect scale, whereas it raised the spiral host by \num{6.73}~pp. %

As at insect scale, the host mattered more than the gate, %
since in the first robot-scale test the plain spiral host reached the source in %
\SI{57.27}{\percent} of episodes and the gated moth model in \SI{48.87}{\percent} (Supplementary %
Table~S29). Unlike at insect scale, the gain held at reversals of \SI{170}{\degree} or more %
(\numrange{23.6}{33.8}~pp over the three hosts and nine flows, post hoc; Supplementary Table~S35), %
possibly because six of those nine flows were square-wave flows on which the plain host often failed %
even at insect scale (Supplementary Table~S22).

\section{Discussion}

Whether a hold gate is worth more than a fixed memory of the earlier wind depends on the setting. In
the rescaled simulation with the robot's sensors, its turn-rate limit and an estimated gait sway,
the gate beat the best fixed delay for both followers tested, although only the spiral host's margin
survived the Bonferroni check, whereas at the moth model's own scale the best delay was not
resolvably worse with either anemometer. Both memories keep the follower on
the wind in which the old odour travelled after a reversal, but they differ in when they use it. A
delay holds every change of the wind for a fixed time, in contact or not, whereas the gate holds the
earlier wind only while the live wind contradicts it and contact is recent. A practical difference
follows, because the best delay changed eightfold between the two anemometers at insect scale and %
fourfold between the two hosts at robot scale, while the gate ran with the same constants %
throughout.

On flows drawn at random, the design rule itself, distrusting a live wind that contradicts the wind
sensed when contact began, raised every follower we tested, five at insect scale and three in the %
rescaled simulation. Its gain tracked how much more often the stored heading than the live wind
pointed to the source when it acted, and this relation held on episodes separate from those that
gave the account. The rule pays where old odour survives a reversal and the host still fails, so in
descriptive and post hoc splits it gave little where reversals were Poisson-timed or nearly
\SI{180}{\degree}. In the latter case an exploratory diagnosis suggests that the plain follower
rarely failed and a small harm from stale stored headings outweighed the rescues. That diagnosis
also suggests refreshing the stored heading when the live wind reverses
during contact, a change we have not tested. The gate also lowered its host where the stored heading
was no better than the live wind, as in the computed house, and it cost the moth model a few points
in steady wind at insect scale.

For a robot builder the gate amounts to one stored angle, a timer since contact and one angle
comparison per step, wrapped around an existing follower. Three results limit when it is worth
adding. First, the gate is an add-on rather than a better controller, since the plain spiral host
beat the gated moth model at insect scale (two-sided, post hoc) and descriptively at robot scale.
Moreover, because no host was tuned to our flows, a tuned host might leave the gate less to
correct. Second, small screens on chosen flows got the sign of the
gate's effect wrong for three hosts, so whether a given host benefits needs a paired test on the
builder's own flows. Third, the host must be able to search at the robot's turn rate, which
surge-cast, whose casting needs fast turns, could not do at the Go1's rate, gated or not.

The evidence for a robot is indirect, because the robot never ran the gate and its trials show only
that a reversal turns the sensed wind against the source over the seconds the arena's air takes to
reverse. No simulation of the arena could specify a hardware test, because with saturated gas old
and new odour are indistinguishable. Such a test therefore first needs a plume model matched to the
arena or measured gas and flow. The rescaled simulation is the insect-scale plume model expressed in
metres, and it departs from the robot in four ways. Its gas sensors respond instantly, without the
rise and recovery of metal-oxide sensors on which the gate's contact onset and \SI{2}{\second} %
discard depend. Its wind turns faster than the arena's air, its gait sway is an estimate, and its
turn cap omits the robot's yaw slew limit. The flows came
from two planar plume models and a distribution we chose rather than from measured environments,
and how often and how far real indoor flows reverse is not known. Finally, the gate's discard time
was set at insect scale with no established scaling, and its conflict angle was not tuned at robot
scale.

Insects also use cues the gate ignores. Flies sense the direction in which odour moves across their
antennae~\cite{kadakia2022odour} and adjust to the timing of odour
encounters~\cite{alvarezsalvado2018elementary,demir2020walking,stupski2024wind}, and such a cue
could tell old odour from new directly, whereas the gate infers it from a disagreement with memory.

In our simulations the structure of the flow and the robot's own sensors and turn rate changed which
memory of the wind paid. Benchmarks used to train or score search
policies~\cite{singh2023emergent,rando2025qlearning,rigolli2022learning,loisy2022searching,heinonen2023optimal}
should therefore include reversing flows of more than one structure as well as the robot's own
sensors and turn rate.

\section{Methods}

\subsection{Robot trials}

\emph{Robot and sensors.} A Unitree Go1 EDU quadruped carried a Raspberry Pi 5 running the
controller under velocity control. A two-dimensional thermal anemometer sat on a mast over the body
centre, about \SI{63}{\centi\metre} above the floor, where we assume, without having measured it,
that the body's wake does not reach it. A left--right pair of MiCS-5524 metal-oxide gas sensors sat
on a 3D-printed front mast. The anemometer updates at \SI{1.96}{\hertz}, one reading every %
\SI{0.51}{\second}, with a circular standard deviation of \SI{21.3}{\degree} at rest in near-still %
air, of \SIrange{3}{7}{\degree} while standing in the arena's fan wind at the trial settings, and of %
\SI{20}{\degree} at the weakest setting. Because its response to %
a change of direction was not measured, every Go1 anemometer model here represents it by a
first-order lag equal to the reading interval, as a proxy. The robot's control software clipped the
commanded yaw rate at \SI{0.5}{\radian\per\second} and limited its rate of change to %
\SI{1.5}{\radian\per\second\squared}. %

\emph{Arena and protocol.} Two fan banks (\SI{40}{\centi\metre} square) faced each other
\SI{2.5}{\metre} apart with no side walls. Under reversal they swapped every \SI{20}{\second}, the %
two directions differing by \SI{146.5}{\degree}, whereas in steady wind only the source-side bank %
blew. An ethanol source sat \SI{45}{\centi\metre} in front of one bank, and the robot started near %
the other, \SI{1.6}{\metre} from the source and facing it. The wind speed is known only from the %
robot's uncalibrated anemometer, which read \SI{1.18}{\metre\per\second} standing at the start. %
Five controllers ran three times in each condition: surge-cast, the moth
model~\cite{yamada2021multisensory}, a reliability-modulated surge of our own inspired by the
repeated surges of moths to successive filaments~\cite{vickers1994reiterative}, posterior-mean
infotaxis (Controllers), and the frozen heading, which is the held heading without re-acquisition.
An overhead camera tracked an ArUco marker for scoring only. A run ended by remote stop when, judged
by eye, the robot had reached the source or left the arena, or otherwise after \SI{90}{\second}. %
The score, settled distance (the mean distance over the last \SI{10}{\percent} of a run), was %
adopted after the trials. Outcomes and closest approach are in Supplementary Table~S3, and the runs
in Supplementary Fig.~S1.

\emph{Sensed-wind cue.} For every control tick, $c=\cos(\theta_u-\beta_s)$, with $\theta_u$ the
anemometer's upwind direction in the world frame and $\beta_s$ the camera bearing from the robot to
the source. The anemometer's heading frame and the camera frame differ by an offset that varied
between runs, so each run was calibrated on its own first \SIrange{3}{15}{\second}, when only the
source-side bank had blown, and those ticks were excluded. The negative cue
\SIrange{10}{20}{\second} after a reversal holds under all three corrections (Supplementary %
Table~S1), which lists the three corrections. Supplementary Table~S2 gives the settling of the
sensed direction after a reversal. Intervals resample runs (20{,}000 resamples).

\subsection{Simulated benchmark}

The source sits at the origin of a planar arena of $x\in[-1100,1100]$ and $y\in[-700,700]$~mm. An %
episode releases the simulated robot at $x=\pm300$~mm with $|y|\le120$~mm and a random heading, and %
succeeds when the body centre comes within the success radius before \SI{300}{\second} %
(Table~\ref{tab:sim}), and leaving the arena is a failure. Unless stated otherwise, the wind
alternates between two directions every \SI{10}{\second}, and every episode warms the plume and wind
under reversal before release and randomizes the reversal phase. The robot is an insect-scale
quadruped with heading drift and bilateral gas sensing, whose gait enters only as a lateral sway, \SI{200}{\milli\metre\per\second} %
at \SI{2}{\hertz}, added to the wind its anemometer reads. Old odour therefore matters more here than
in the fan arena, because the ratio of the odour's travel time to the interval between reversals is
about \num{0.19} in the benchmark against \numrange{0.04}{0.07} in the arena. %

\begin{table}[h]
\caption{Benchmark constants.}
\label{tab:sim}
{\footnotesize
\begin{tabular}{@{}llr@{}}
\toprule
benchmark constant & symbol & value \\
\midrule
wind speed & $U$ & \SI{160}{\milli\metre\per\second} \\
anemometer lag & $\tau_w$ & \SI{0.10}{\second} \\
anemometer noise (s.d.) & $s_w$ & \SI{5.0}{\degree} \\
heading drift & $\dot{h}_d$ & \SI{4}{\degree} per $\sqrt{\mathrm{s}}$ \\
episode cap & $T_{\max}$ & \SI{300}{\second} \\
success radius (distance to source) & $r_g$ & \SI{50}{\milli\metre} \\
antenna offset from body centre & $d_A$ & \SI{13.23}{\milli\metre} \\
antenna half-angle & $\alpha_A$ & \SI{19}{\degree} \\
gait sway, lateral velocity amplitude & $a_g$ & \SI{200}{\milli\metre\per\second} \\
gait sway frequency & $f_g$ & \SI{2}{\hertz} \\
\midrule
\multicolumn{3}{@{}p{0.92\linewidth}}{\footnotesize The simulated platform is an insect-scale quadruped: an FT602 anemometer model with IMU drift and two antennae at $\pm\alpha_A$. Its gait is modelled only as a lateral body sway of amplitude $a_g$ at $f_g$, added to the apparent wind the anemometer reads and not to the path; it is not fitted to or validated against the Go1. A detection is either antenna's gain-scaled concentration crossing the fixed threshold. An episode succeeds when the body centre comes within $r_g$ of the source before $T_{\max}$. These are the \emph{benchmark's} constants; the physical rig's anemometer differs and is quoted only where the rig is discussed.} \\
\bottomrule
\end{tabular}}

\end{table}

\emph{Plume models and anemometer.} Pompy~\cite{farrell2002filament}
(\url{https://github.com/InsectRobotics/pompy}) moves filaments with a \num{15}$\times$\num{11}
stochastic grid field that relaxes toward the forcing wind and receives coloured noise at its
boundary; the field diffusivity $k$ (default \num{20}) sets how quickly the field at the robot
follows a reversal, from a median \SI{3.6}{\second} at the default to \SI{0.2}{\second} at $k=320$, %
and the boundary-noise gain (default \num{2}) its spatial irregularity. The forcing wind switches at
once at a reversal. The anemometer reads the field velocity at the robot through apparent-wind,
gait, lag (\SI{0.1}{\second}) and noise (\SI{5}{\degree}) models of an ultrasonic anemometer, with %
heading drift \SI{4}{\degree} per $\sqrt{\mathrm{s}}$. The puff model moves discrete puffs with the
forcing wind. GADEN~\cite{monroy2017gaden,ojeda2025gadenrt} ran a computed house of the VGR
dataset~\cite{vgr_dataset}, switching every \SI{10}{\second} between two steady ventilation solutions
(simpleFoam, $k$--$\omega$ SST) at once or blended over \SI{2}{\second} or \SI{5}{\second}. For the
house solved through the switch we ran its mesh transiently with OpenFOAM~\cite{weller1998openfoam}
(pimpleFoam), alternating the two boundary-condition sets every half-period of 10, 20 or
\SI{40}{\second} of benchmark time for 20 cycles, until the last two cycles differed by at most %
\num{0.0092} in speed-weighted median relative difference (Supplementary Table~S24). %

\emph{Flows.} The 31 chosen flows vary pompy's irregularity and relaxation, the reversal schedule
(every \SI{2.5}{\second} or \SI{20}{\second}, Poisson-timed with a mean of \SI{5}{\second}, a
\SI{120}{\degree} separation, a \SI{900}{\milli\metre} release) and add the puff model and the
computed house (Supplementary Table~S5). The 60 random flows were drawn once from a fixed generator seed:
boundary-noise gain uniform on $[0,2]$, diffusivity log-uniform on $[20,320]$, separation uniform on %
$[\SI{120}{\degree},\SI{180}{\degree}]$, square or Poisson schedule with equal probability, mean
reversal interval log-uniform on \SIrange{2.5}{20}{\second}, and release range uniform on
\SIrange{300}{600}{\milli\metre} (Supplementary Table~S36). No flow was replaced or dropped. Robot-scale test %
flows were drawn from the same distribution, with the release range scaled, under generator seeds no
screen used. Steady-wind episodes removed the reversal and released the robot downwind on the test
flows' fields.

\subsection{Controllers}

\emph{Followers.} The wind followers surge toward the sensed upwind on contact at %
\SIrange{18}{22}{\milli\metre\per\second} and differ in how they search after losing it. Surge-cast %
zigzags about upwind at a fixed amplitude after \SI{1.5}{\second} without contact. The moth %
model~\cite{yamada2021multisensory} surges, zigzags with a speed set by the recent detection
frequency, then loops, at the published constants (Table~\ref{tab:params}). SZL ports the silkmoth's
surge-zigzag-loop program~\cite{kanzaki1992zigzag}. The spiral host runs an expanding Archimedean
spiral after \SI{1}{\second} without contact. The count-gated surge surges only after two detections %
within \SI{10}{\second}. Infotaxis~\cite{vergassola2007infotaxis} keeps a belief over the source
position from a hit-rate model; in simulation it steps to the neighbouring position that most reduces
the expected entropy, and on the robot it steered toward the posterior mean of the last
\SI{22}{\second} of records. Every follower ran at fixed constants (Supplementary Table~S6). %

\emph{Hold gate.} The gate pre-processes the follower's wind input (Algorithm~\ref{alg:gate},
Table~\ref{tab:params}). A detection counts only while the robot heads within $\theta_d$ of the sensed
upwind. The stored heading is taken only when none is stored, so it is not refreshed during
uninterrupted contact, and it is discarded after $t_a$ without contact. The same code wraps every
host. The constants were set on this benchmark when the held heading was developed. When they were
varied one at a time on the chosen flows, conflict angles of \SI{105}{\degree} and \SI{120}{\degree} scored %
\numrange{2.2}{2.9}~pp above the \SI{90}{\degree} used (exploratory; Supplementary Table~S37). %

\begin{algorithm}[h]
\caption{One control step of the hold gate around a wind follower.}
\label{alg:gate}
\begin{algorithmic}[1]
\State \textbf{inputs} observation $o$ with sensed upwind $\theta_u$, heading $h$, detection bits $(d_L,d_R)$, timestep $\delta$
\State \textbf{state} stored heading $b$ (initially unset), time since contact $\tau=0$
\State $c\gets(d_L\lor d_R)\wedge|\mathrm{wrap}(h-\theta_u)|\le\theta_d$ \Comment{gate's own contact test; $o$'s detection bits are not changed}
\If{$c$}
    \State $\tau\gets0$; \textbf{if} $b$ is unset \textbf{then} $b\gets\theta_u$ \Comment{acquire; not refreshed in contact}
\Else
    \State $\tau\gets\tau+\delta$; \textbf{if} $\tau>t_a$ \textbf{then} $b\gets\textsc{unset}$ \Comment{discard}
\EndIf
\If{$b$ is set \textbf{and} $|\mathrm{wrap}(\theta_u-b)|>\gamma$}
    \State $o.\theta_u\gets b$ \Comment{the follower sees the stored heading; nothing else in $o$ changes}
\EndIf
\State \Return $\textsc{Follower}(o)$
\end{algorithmic}
\end{algorithm}

\begin{table}[h]
\caption{Constants of the hold gate and of its first host, the moth model.}
\label{tab:params}
{\footnotesize
\begin{tabular}{@{}lllr@{}}
\toprule
 & Symbol & Meaning & Value \\
\midrule
Hold gate & $t_a$ & discard after this long without gated contact & 2 s \\
 & $\gamma$ & conflict threshold between live wind and stored heading & 90$^\circ$ \\
 & $\theta_d$ & detections count only within this angle of upwind & 120$^\circ$ \\
Moth model & $v_s$, $v_c$ & surge and cover speed & 22, 14 mm/s \\
 & $t_s$ & straight surge after a detection & 0.5 s \\
 & $A_z$ & zigzag amplitude about upwind & 1 rad \\
 & $t_l$ & loop after this long without detection & 5.7 s \\
 & $\omega_l$ & loop turn rate & 1.5 rad/s \\
 & $k_v$ & speed gain on detection frequency & 8 mm \\
 & $\theta_h$ & host's detection gate & 90$^\circ$ \\
\bottomrule
\end{tabular}}

\end{table}

\emph{Choosing the gated controller.} The host and the form of the gate were chosen on a selection
block (60 episodes per flow) disjoint from every test block. On eight chosen flows the gate as adopted %
reached \SI{60.4}{\percent} and the plain moth model \SI{57.3}{\percent}. The alternative forms did %
worse, since a stored heading that tracks the live wind while they agree, with or without a limit on
the override, reached \SIrange{55.6}{56.5}{\percent}. The gate also lowered surge-cast from \SI{57.3}{\percent} to %
\SI{47.5}{\percent}. A screen of four further hosts on eleven chosen flows (40 selection episodes %
each; Supplementary Table~S38) showed the gate raising the count-gated surge and infotaxis and
lowering the spiral host and SZL. We then tested the count-gated surge and SZL, and later surge-cast
and the spiral host, on the random flows.

\emph{Held heading.} The held heading (Algorithm~\ref{alg:wr}, Table~\ref{tab:params_wr}) stores the
sensed upwind at its first gated detection, discards it after $t_a$ without gated contact, rejects a
live wind that disagrees with it by more than $\gamma$, and between contacts casts about the stored
heading. Re-acquiring a new heading at the next contact raised its success on the main benchmark from
\SI{21.0}{\percent} to \SI{58.25}{\percent} (Supplementary Table~S39). %

\begin{algorithm}[h]
\caption{One control step of the held heading.}
\label{alg:wr}
\begin{algorithmic}[1]
\State \textbf{inputs} gas intensities $(g_L,g_R)$, detection bits $(d_L,d_R)$,
       sensed upwind $\theta_u$, heading $h$, timestep $\delta$
\State \textbf{initial state} $b$ unset; $r=s=\tau_c=0$, $q=1$, $\tau=\infty$
\If{$|\mathrm{wrap}(h-\theta_u)|>\theta_d$} $d_L\gets0$; $d_R\gets0$
\EndIf
\State $\kappa\gets1$ if both detections, $w_1$ if exactly one, else $0$
\State $r \gets r + \beta\,(\kappa-r)$ \Comment{reliability EWMA}
\If{$s = 0$ \textbf{and} $r > r^{+}$} $s \gets 1$
\ElsIf{$s = 1$ \textbf{and} $r < r^{-}$} $s \gets 0$
\EndIf \Comment{hysteresis on $r^{\pm}$}
\State $\tau\gets0$ if $d_L\lor d_R$, else $\tau+\delta$
\If{$b$ is set \textbf{and} $\tau > t_a$}
    \State $b\gets\textsc{unset}$ \Comment{discard after $t_a$ without gated contact}
\EndIf
\State $\textit{conflict}\gets(b\text{ is set})\wedge|\mathrm{wrap}(\theta_u-b)|>\gamma$
\State $\textit{trust}\gets(s=1)\wedge(d_L\lor d_R)\wedge\neg\textit{conflict}$
\If{$b$ is unset \textbf{and} $(d_L\lor d_R)$}
    \State $b\gets\mathrm{wrap}(\theta_u)$ \Comment{acquire at the first gated contact}
\EndIf
\State $C\gets0$
\If{$b$ is unset}
    \State $\theta^*\gets\theta_u$; $v\gets v_b$ \Comment{bootstrap upwind}
\ElsIf{$\textit{trust}$}
    \State $\theta^*\gets\theta_u$; $v\gets v_s$
    \State $C\gets k_g(g_R-g_L)/(g_R+g_L+\epsilon)$
\Else
    \State $\tau_c\gets \tau_c+\delta$
    \If{$\tau_c\ge T_c$} $q\gets-q$; $\tau_c\gets0$
    \EndIf
    \State $\theta^{*} \gets \theta_u$ if $\tau < t_r$ and $\neg\,\textit{conflict}$, else $b$
    \State $\theta^*\gets\mathrm{wrap}(\theta^*+qA_c)$; $v\gets v_a$ \Comment{alternate cast offsets}
\EndIf
\State $\omega\gets k_\theta\,\mathrm{wrap}(\theta^*-h)-C$
\State \Return $(\operatorname{clip}(v,0,v_{\max}),
       \operatorname{clip}(\omega,-\omega_{\max},\omega_{\max}))$
\end{algorithmic}
\end{algorithm}

\begin{table}[h]
\caption{Parameters of the held heading.}
\label{tab:params_wr}
{\footnotesize
\begin{tabular}{@{}lp{0.50\linewidth}r@{}}
\toprule
symbol & quantity & value \\
\midrule
$\delta$ & control timestep & \SI[parse-numbers=false]{1/30}{\second} \\
$\beta$ & reliability EWMA rate & \num{0.08} \\
$w_1$ & single-antenna weight in the coincidence signal & \num{0.40} \\
$\theta_d$ & detection gate half-angle & \SI{120}{\degree} \\
$r^{\pm}$ & reliability switch thresholds (hysteresis) & \num{0.16} $\pm$ \num{0.06} \\
$\gamma$ & conflict threshold & \SI{90}{\degree} \\
$t_a$ & discard timeout since gated detection & \SI{2.0}{\second} \\
$t_r$ & recency window for the adaptive cast target & \SI{3.0}{\second} \\
$A_c$ & cast angular offset & \SI{0.9}{\radian} \\
$T_c$ & cast half-period (time spent casting) & \SI{1.0}{\second} \\
$k_\theta$ & heading gain & \num{3.0} \\
$k_g$ & relative bilateral centering gain & \num{2.2} \\
$\epsilon$ & centering denominator offset & \num{1e-06} \\
$v_s$ & surge speed & \SI{22.0}{\milli\metre\per\second} \\
$v_a$ & re-acquire speed & \SI{20.0}{\milli\metre\per\second} \\
$v_b$ & bootstrap speed before the first commitment & \SI{18.0}{\milli\metre\per\second} \\
$v_{\max}$ & speed limit & \SI{40.0}{\milli\metre\per\second} \\
$\omega_{\max}$ & yaw-rate limit & \SI{8.0}{\radian\per\second} \\
\midrule
\multicolumn{3}{@{}p{0.92\linewidth}}{\footnotesize The controller is deterministic given its inputs. The discard timeout $t_a$ is selected per arena; no transferable scaling has been established.} \\
\bottomrule
\end{tabular}}

\end{table}

\emph{Wind-history controllers.} The averaging controller surges toward the circular mean of the
sensed upwind over its window while a detection is recent (its surge hold) and otherwise casts about
that mean. The source-likelihood controller, after ref.~\cite{pang2006chemical}, traces the sensed
wind back from each detection over ten odour ages up to its horizon, adds the traced positions to a
grid map with a Gaussian spread of \SI{80}{\milli\metre} and steers to the map's peak. The tuning %
screen crossed windows of 2 to \SI{60}{\second} with surge holds of 1.5 or \SI{5}{\second}, and %
horizons of 3, 10 or \SI{30}{\second} with a map decay, $\exp(-\delta/\tau)$ per tick with $\tau$ of %
3, 10 or \SI{30}{\second} or none; the published method keeps all evidence, and the decay is our %
addition. Each controller took its configuration of highest mean success on the screen
(Supplementary Table~S7); on 13 chosen flows both had run at their defaults (Supplementary Table~S40). %

\emph{Low-pass and delay.} The low-passed follower steers by a first-order circular low-pass of the
sensed upwind; each tick the filtered heading moves toward the reading by $\delta/\tau$ of their
wrapped difference. The delayed follower steers by the sensed upwind of
$K=\lfloor L/\delta+\tfrac12\rfloor$ control ticks earlier, $\delta=\SI[parse-numbers=false]{1/30}{\second}$, %
with detections, pose and time live; with $L=0$ it reproduced the moth model's outcomes exactly. In
the gated forms the filter or delay sits before the gate, which stores, compares and substitutes the
filtered or delayed reading, while contacts stay live. Screens chose the delay from 0.25, 0.5, 1, 2 %
and \SI{4}{\second}, per anemometer at insect scale and per host at robot scale, and the low-pass
time constant from 0.5 to \SI{8}{\second} at insect scale (from 0.25 to \SI{4}{\second} in the %
low-pass test), by highest mean success, ties to the smaller (Supplementary Table~S8; Supplementary Table~S34).

\subsection{Robot-scale simulation}

The robot-scale benchmark maps the insect-scale one by a similarity: every length is multiplied by
$\lambda=\num{7.5}$ and time is unchanged, so the \SI{160}{\milli\metre\per\second} wind becomes %
\SI{1.2}{\metre\per\second} and every reversal interval, sensor constant and controller time %
constant keeps its value in seconds. The plume model runs in its own units at $\lambda$ times the
length per unit, so its field and filaments are those of the insect-scale episode with the same
index, and controllers receive positions divided by $\lambda$ and have their speed commands
multiplied by $\lambda$. With
the benchmark's sensors the map reproduced 75 episodes each of the plain and gated moth model exactly %
in outcome and in time to \SI{0.05}{\second}. The simulated robot then took the Go1's anemometer %
model (lag proxy, noise and \SI{1.96}{\hertz} sample-and-hold), a heading drift of %
\SI{1}{\degree} per $\sqrt{\mathrm{s}}$, two gas sensors \SI{40}{\milli\metre} from the body centre %
at $\pm\SI{45}{\degree}$, and walls of a \SI{16.5}{\metre} by \SI{10.5}{\metre} hall that stop it. A %
plume probe without a controller found the concentration's crosswind spread $\sigma_y$ to be
\numrange{0.17}{0.23} of the distance. Near a source, Taylor's theory makes $\sigma_y/d$ about the %
turbulence intensity~\cite{taylor1922diffusion}, so this spread implies a turbulence intensity that
we did not measure in a real hall
(Supplementary Table~S28).

\emph{Settings.} The first setting kept the controllers' \SI{8}{\radian\per\second} turn-rate limit %
and the \SI{200}{\milli\metre\per\second} gait sway of a rig-matched simulation, an assumed value; %
its noise was the \SI{21.3}{\degree} of near-still air. The second clipped each commanded turn rate %
at the robot's \SI{0.5}{\radian\per\second} by an outermost wrapper of which the host and gate knew %
nothing, without the slew limit, and raised the sway to \SI{410}{\milli\metre\per\second}. That %
value reproduces, through the simulated anemometer's lag, the direction wobble that walking added
over standing in the robot's logs, \SI{2.1}{\degree} (95\% bootstrap interval \numrange{0}{5.1} over %
runs). Because the logs do not resolve the gait, this value is an estimate. Without the lag the same
wobble needs only about \SI{63}{\milli\metre\per\second}, and the camera bounds the body's lateral %
velocity at about \SI{54}{\milli\metre\per\second}, so the estimate errs high if the real sensor %
follows a \SI{2}{\hertz} sway more closely than the lag model. The third set the anemometer noise to %
\SI{6.8}{\degree} per reading, the standing value under the source-side bank at its trial setting. %
This is the larger of the two in-range sweep values and lies above the median, \SI{4.26}{\degree}, %
of seven standing windows in operating wind. The paired comparison ran all three and a fifth setting, the
first with only the noise lowered, on the third test's 60 flows and episode indices. At insect scale,
the Go1 anemometer model (\SI{21.3}{\degree} noise, lag, hold, drift) with the %
\SI{200}{\milli\metre\per\second} sway ran on 60 new random flows and, in the delay test, on the %
random flows; on the six diagnosis flows the benchmark's anemometer was changed one property at a
time (Supplementary Table~S25). A simulation of the fan arena itself (rig-matched; Supplementary Table~S26; Supplementary Table~S27) used the
arena's geometry and \SI{146.5}{\degree} reversals with saturated gas or a pompy plume. %

\subsection{Diagnostics}

\emph{Probe and outdoor data.} Thirty stationary points at 150, 300 and \SI{450}{\milli\metre} from %
the source on ten bearings were sampled every tick for \SI{40}{\second}, 60 episodes per flow; a
detection is the robot's test, and upwind is the sensor's reading at the point without lag or noise.
We reanalysed the odour and wind records of Nag and van Breugel~\cite{nag2024odour}, computing
$c=\cos(\theta_u-\beta_s)$ for every odour sample from the dataset's positions. A wind shift is a
change of the smoothed upwind by more than \SI{90}{\degree} within \SI{10}{\second}, of which the %
forest records contain 44, and intervals come from a moving-block bootstrap over \SI{300}{\second} blocks (descriptive; Supplementary Table~S4). %

\emph{State-resolved diagnosis and the account.} On six diagnosis flows (the puff model, two
noise-free pompy fields, the \SI{120}{\degree} reversal, the default flow and reversals every
\SI{2.5}{\second}), and for the gate on five more (two flows where it had trailed the best controller
and the three house variants), we logged every tick of 40 selection episodes per controller. Each
tick was labelled by contact and by the sign of the cosine between sensed upwind and the bearing to
the source. At each override, the stored heading and the live upwind were scored as on the source's
side when their cosine with the bearing was positive. The account is the override share times the
difference between the two scores. In the host test it came from the test's own gated episodes; the
cross-fitted version split each cell's episodes by index parity and averaged the Spearman correlations
of the account on one half with the gain on the other, in both directions (Supplementary Table~S41).

\emph{Near-full reversal diagnosis.} We reran the plain and gated moth model with every tick logged,
50 paired episodes per cell on a fresh block, at insect scale and with the first robot-scale %
setting, on the random flows of \SI{170}{\degree} or more at their own separation and at
\SI{135}{\degree}, on the robot-scale test's flows of \SI{170}{\degree} or more, and on eleven random
flows of \SIrange{120}{150}{\degree} set to \SI{175}{\degree}. On paired episodes the gain is
$r(1-p)-h\,p$, with $p$ the host's success, $r$ the share of its failures the gated form solved
(rescue) and $h$ the share of its successes the gated form failed (harm). A stored heading counts as
stale when the true wind jumped by more than \SI{60}{\degree} since it was taken. Intervals resample %
flows (4{,}000 times); the diagnosis selected nothing and reports no tests. %

\subsection{Statistics and reproducibility}

Every simulated episode is a deterministic function of its integer index, which sets the release
pose, heading, reversal phase and every random seed; flows set only the wind and plume parameters.
Controllers run on the same indices are therefore paired exactly, and within a block every flow
reuses the same indices, so flows and episode indices are crossed. Indices come from fixed seed
blocks that the released code checks to be pairwise disjoint; selection and testing never share a
block, and each test has its own block, except the paired robot-scale comparison, which by design
reran one test's flows and indices. Twelve tests compared the gate with its host or with other
controllers, or its gain between settings, each with controllers and constants fixed on selection or
screening blocks; Supplementary Table~S42 lists each test's setting, flows, episode block, episodes per flow and
main comparisons, 45 in all. %

\emph{Intervals and P values.} An effect is the mean over flows of the per-flow paired difference in
success, in pp. For every main comparison, the 95\% interval is a two-sided percentile interval from
20{,}000 crossed bootstrap resamples (\texttt{default\_rng(0)}) that draw flows and episode indices
independently with replacement and average the difference over the drawn flows and indices; this
respects the crossing of flows and indices. A one-sided bootstrap $P$ for a gain is
$(1+n_{\le0})/(B+1)$, where $n_{\le0}$ counts resamples at or below zero; a two-sided $P$ is
$\min(1, 2\min(1+n_{\le0},1+n_{\ge0})/(B+1))$. $P$ values are one-sided, for a gain of the first-named
controller, unless marked two-sided; the paired robot-scale comparison and the robot-scale delay test
were designed two-sided. The smallest attainable one-sided $P$ is $1/20{,}001\approx\num{5e-5}$, %
reported as a bound. Within each test, $P$ values are Holm-adjusted over its main comparisons
(post hoc for the five earliest tests, which were not designed with a Holm family) and judged at
$\alpha=0.05$. As a stricter post hoc check we applied a Bonferroni correction over all 45 main %
comparisons, judging each interval comparison by its crossed-bootstrap tail in the direction of its
estimate against $0.05/90$ and each correlation by its $P$ against $0.05/45$. Of the 31 comparisons %
whose 95\% interval excludes zero, 18 pass, with tails at most \num{5.5e-4}, and 13 fail, with tails %
from \num{8.5e-4} to \num{0.023}. With four of the six correlations also passing, 22 of the 45 pass %
in all (Supplementary Table~S43). %

The tests were planned with intervals that resampled flows and then episodes within each drawn flow,
which ignores the crossing of flows and indices. The crossed intervals were %
\numrange{1.04}{1.79} times as wide (median \num{1.19}), and six to seven times as wide for the %
chosen-flow test, whose planned intervals resampled only episodes; Supplementary Table~S43 gives
both for every main comparison. We had decided to keep the planned intervals only if no comparison
changed its relation to zero and no interval widened by more than about a quarter, and because that
rule failed, every main comparison in this paper is reported with the crossed bootstrap. Under it
two comparisons no longer exclude zero, the chosen-flow margin over surge-cast and the gain from
gating the delayed follower with the Go1 anemometer at insect scale, while five more lose
significance after Holm adjustment and four more fail the Bonferroni check (Supplementary
Table~S43). Descriptive and post hoc intervals (splits, per-setting gains and the diagnosis) still
resample flows and then episodes, and so probably understate their width by a similar margin.

\emph{Correlations.} Correlations of gain and account are Spearman's $\rho$ over flow-by-host cells,
tested one-sided for $\rho>0$ by permuting cells (20{,}000 permutations). Because cells of one flow
share its wind field and seeds, we also permuted whole flows, keeping each flow's gains in their host
slots and pairing them with another flow's accounts (Supplementary Table~S20; post hoc). For the host test
we also computed a partial correlation given separation, schedule, the plain host's success and the
host (post hoc). Per-flow comparisons use exact two-sided McNemar tests, Holm-corrected over the
flows of a test, single rates carry Wilson intervals, and the robot's statistics are descriptive.
Success levels pair only within a block, because a block's draw of release poses and seeds is shared
by all its flows (Supplementary Table~S44). Exploratory analyses (the screens, the anemometer models, the solved
house, the arena simulations and the near-full diagnosis) are reported without correction.

\subsection{Use of AI tools}

AI tools were used to help write code, preprocess data and draft and edit the manuscript text.
All research decisions and the direction of the study were made by the authors, who checked
every statement and take full responsibility for the content.

\subsection{Code availability}

The simulation code, the controllers, the robot deployment code, the transient house case, the
analysis plans and scripts, the seed blocks and the code that emits every figure and table from
the run records are in a repository whose link is provided to the editors and referees at
submission, and will be made public at publication. Every simulated episode, and so every
simulation result, regenerates from this code and its seed blocks.

\backmatter

\bmhead{Supplementary information}
Supplementary Tables S1--S44 and Supplementary Fig.~S1. Supplementary Video 1 shows %
steady-wind and reversing-wind runs of the frozen heading and of the surge-cast controller on the Go1
from the overhead scoring camera, each the run with the median settled distance of its three, at
three times real time.

\bmhead{Data availability}
The per-episode simulation records for every number in this paper, the analysis outputs, the
seed blocks, the sampled velocity fields of the solved house and the thirty robot runs
(controller logs, camera tracks and videos) are in the same repository, whose link is provided to
the editors and referees at submission, and will be deposited on Zenodo with the code on publication.

\bmhead{Acknowledgements}
This work was supported in part by JSPS KAKENHI Grant Number JP25H01157.

\bmhead{Author contributions}
J.S. and S.S. conceived the study; S.S. supervised the work and provided the robot and facilities
at NII; J.S. built the controllers and simulations, ran the experiments and analysis, and wrote
the manuscript.

\bmhead{Competing interests}
All authors declare no financial or non-financial competing interests.

\bibliography{references}

\end{document}